\documentclass{article}

\usepackage[main,final]{neurips_2026}

\usepackage[utf8]{inputenc} %
\usepackage[T1]{fontenc}    %
\usepackage{hyperref}       %
\usepackage{url}            %
\usepackage{booktabs}       %
\usepackage{multirow}       %
\usepackage{colortbl}       %
\usepackage{amsmath}        %
\usepackage{amsfonts}       %
\usepackage{amssymb}        %
\usepackage{nicefrac}       %
\usepackage{microtype}      %
\usepackage{xcolor}         %
\usepackage{graphicx}
\usepackage{subfigure}
\usepackage{fancyvrb}
\usepackage{tikz}
\usepackage{pgfplots}
\pgfplotsset{compat=1.18}

\title{FrameRouter: Frame Budget Routing for Long-Video Understanding on Video-MME-v2 and Beyond}

\author{%
  Bingjun Luo \\
  Tsinghua University \\
  Beijing, China \\
  \texttt{bingjunluo@outlook.com}
  \And
  Jialin Guo \\
  Harbin Engineering University \\
  Harbin, China \\
  \texttt{guojialin@hrbeu.edu.cn}
  \And
  Siqi Li\thanks{Corresponding author: Siqi Li (\texttt{lisiqi19971013@gmail.com}).} \\
  BNRist, THUIBCS, School of Software \\
  Tsinghua University \\
  Beijing, China \\
  \texttt{lisiqi19971013@gmail.com}
}

\begin{document}

\maketitle

\begin{abstract}
  Long-video question answering benchmarks like Video-MME-v2 typically evaluate systems under a uniform frame budget for every question, despite stark variation in the evidence each query actually demands. We analyze this mismatch on Video-MME-v2 and find that different question subsets prefer different frozen frame-budget policies, while uniformly increasing the frame count does not reliably improve all metrics. These findings motivate fixed-mean frame-budget routing, a controlled inference setting in which each question may receive a different frame budget while the average budget matches a uniform reference. We introduce FrameRouter, a training-free test-time framework that combines an evidence-demand router, a budget-constrained allocation rule, and a frozen bank of frame-sampling policies. By controlling \emph{how much} visual evidence each question receives while leaving the underlying samplers unchanged, FrameRouter is orthogonal to query-aware frame selection. Across Video-MME-v2 and several long-video understanding benchmarks, FrameRouter improves long-video question answering by using the same average visual budget more selectively across questions. Code is available at \url{https://github.com/bingjunluo/FrameRouter}.
\end{abstract}

\section{Introduction}

Long-video question answering has become a stress test for multimodal models:
the system must search, aggregate, and reason over visual evidence spread
across minutes or hours of video. Recent benchmarks such as Video-MME-v2,
MLVU, and LongVideoBench make this pressure explicit by evaluating models
across long durations, diverse domains, subtitles, and multi-step question
groups~\citep{fu2026video,zhou2025mlvu,wu2024longvideobench}.
Because compute and context length restrict the number of frames that current
video--language models can ingest, frame budget has become a central evaluation
knob. The de facto protocol is to fix a uniform per-question budget and apply
it to every instance. This is simple and reproducible, but it tacitly assumes
that every question deserves the same amount of visual evidence.

That assumption sits uneasily with the structure of modern long-video
benchmarks. Their questions span heterogeneous evidence demands: some are
resolvable from sparse global context, scene-level commonsense, or subtitle
cues, while others hinge on local visual details, long-range temporal
association, action ordering, or repeated-action counting. Spending the same
number of frames on every question therefore creates two coupled
inefficiencies: low-demand questions consume frames they may not need, and
high-demand questions remain under-observed. Recent efficient video-QA work
makes a related observation, showing that uniformly sampled long videos
contain substantial redundancy and that question-aware frame selection can
recover crucial spatiotemporal cues~\citep{park2026too,zhang2025q,
yao2025generative,ben2026himu}. Our focus is orthogonal to which frames a
question keeps: we ask how much evidence each question should receive under
the same average cost.

Our diagnostic analysis on Video-MME-v2 reinforces this view. Within a frozen
bank of low-, mid-, and high-budget full-video uniform policies, neighboring
budgets repair a non-trivial slice of mid-budget errors while also breaking
questions that the mid-budget policy answers correctly. The three policies do
not form nested error sets; their disagreement varies systematically with
question group and difficulty. This motivates \emph{fixed-mean frame-budget
routing}: given a uniform reference budget $B_{\mathrm{ref}}$ over $N$
questions, an admissible router may assign per-question budgets $B_i$ subject
to $\frac{1}{N}\sum_{i=1}^{N} B_i = B_{\mathrm{ref}}$. Under this strict
equality, any gain must come from reallocating visual evidence across
questions, not from increasing the average number of frames. This framing is
complementary to query-aware frame selectors and to architecture-level token
compressors, which adapt what or how to sample within a fixed per-question
quota; here, we ask how the dataset-level budget itself should be distributed.

\begin{figure*}[t]
  \centering
  \includegraphics[width=0.98\textwidth]{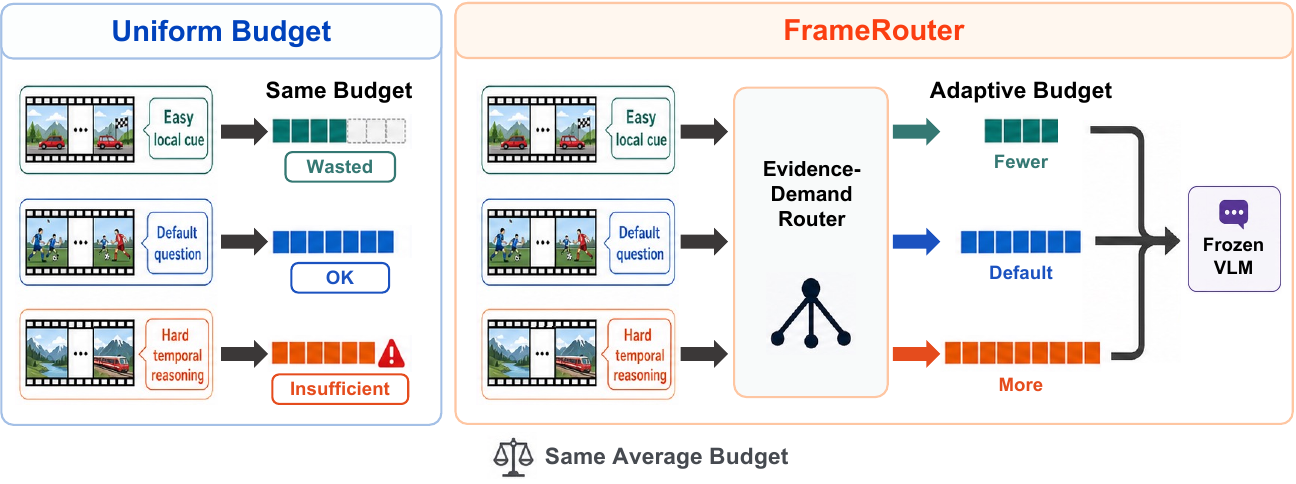}
  \caption{Overview of fixed-mean frame-budget routing. Uniform inference gives
  every question the same visual budget despite heterogeneous evidence demands.
  FrameRouter routes each question to an appropriate frozen budget policy using
  question-only signals, while preserving the same dataset-level average
  budget.}
  \label{fig:intro}
\end{figure*}

We instantiate this formulation as \textbf{FrameRouter}, a training-free
test-time framework that routes \emph{frame budgets}, not frame locations.
FrameRouter scores the benchmark question prompt, whose multiple-choice
options are visible to any evaluated system, and uses those question-only
priorities to choose among frozen low-, mid-, and high-budget full-video
policies under an exact dataset-level mean-budget constraint. The router never
sees the video, subtitles, sibling questions, other videos, labels, or
candidate policy outputs, and it neither fine-tunes the underlying VLM nor
introduces a new within-question frame selector. This keeps the comparison
tightly controlled: low- and high-budget assignments change the distribution
of visual evidence across questions, while the average number of frames remains
identical to the uniform reference. Across Video-MME-v2, LongVideoBench,
MVBench, and VideoMMMU, FrameRouter yields stable matched-budget improvements.
Quota sweeps and post-hoc route audits further show that the benefit is not
simply a high-budget effect, but comes from reallocating the fixed average
budget across heterogeneous questions.

In summary, our contributions are threefold as follows:
\begin{itemize}
  \item We diagnose non-monotonic frame-budget effects on Video-MME-v2 and
  formalize fixed-mean frame-budget routing as a controlled inference problem,
  separating dataset-level budget reallocation from variable-frame efficiency
  and from within-question frame selection.
  \item We instantiate this problem as FrameRouter, combining a frozen
  full-video uniform policy bank, a question-only evidence-demand priority
  parser, and a quota-constrained allocator that holds the mean frame budget
  exactly equal to the uniform reference.
  \item We achieve the state-of-the-art open-source result on
Video-MME-v2 under the matched-budget protocol, and show consistent
improvements across other benchmarks, including LongVideoBench, MVBench, and VideoMMMU.
\end{itemize}

\section{Related work}
\subsection{Long-video understanding benchmarks}
Long-video understanding has moved from short, isolated clips toward benchmarks
that require models to integrate evidence over minutes or hours. EgoSchema
introduced a diagnostic multiple-choice setting over three-minute egocentric
clips, emphasizing that temporal difficulty is not captured by clip length
alone~\citep{mangalam2023egoschema}. Video-MME broadened evaluation across video
duration, domain, and modality, including subtitles and audio, and exposed a
large gap between frontier proprietary systems and open models on long-video
analysis~\citep{fu2025video}. MLVU and LongVideoBench further expanded the
evaluation surface with multi-task long-video understanding and interleaved
video-language contexts~\citep{zhou2025mlvu,wu2024longvideobench}. These
benchmarks made frame budget a practical evaluation variable: model quality
depends not only on the backbone, but also on how many frames or visual tokens
are admitted into the context.

Video-MME-v2 sharpens this pressure by changing both the data distribution and
the scoring rule~\citep{fu2026video}. It contains 800 recent videos and
3,200 questions, organized into groups of four interrelated questions. Beyond
average per-question accuracy, the benchmark reports group-level non-linear
scores, including Non-Lin, Consistency, and Coherence, that reward models for
answering related questions consistently and for preserving multi-step reasoning
chains within a group. These metrics matter for budgeted inference because the
same video can contain questions with very different evidence demands: global
story context, long-range temporal aggregation, local perceptual details, OCR,
or screen evidence. A uniform frame budget is therefore poorly matched to the
heterogeneity that Video-MME-v2 is well suited to surfacing.

\subsection{Efficient video question answering}
The high cost of video QA has motivated methods that reduce redundant visual
processing. Early long-video QA systems often uniformly sampled many frames and
converted them into captions or visual tokens, but long videos contain large
regions that are irrelevant to a particular question. As models and benchmarks
move to longer inputs, evaluation quality depends not only on the VLM backbone
but also on frame budget, sampling density, and whether auxiliary channels such
as subtitles or interleaved text are admitted into the context. The official
64-frame-with-subtitle setting in Video-MME-v2 makes this cost--accuracy
tradeoff explicit: more evidence can help difficult questions, but increasing
the per-question budget uniformly also spends frames on questions that may not
need them.

Existing efficient video-QA methods mostly reduce the cost of processing each
question. LVNet explicitly argues that many sampled frames are unnecessary and
uses a hierarchical keyframe selector to reduce the captioning burden for
long-form QA~\citep{park2026too}. Agentic approaches such as VideoAgent ask an
LLM controller to iteratively search for useful visual information under a
limited frame budget~\citep{wang2024videoagent}. Architecture-level approaches
also attack the cost problem by compressing the video representation: MA-LMM
maintains a memory bank for long-term video information~\citep{he2024ma},
while LongVU uses spatiotemporal adaptive compression to reduce token redundancy
while preserving salient visual detail~\citep{shen2024longvu}. These works
establish that long-video QA is governed by a cost--accuracy tradeoff, but their
primary unit of adaptation is usually the video, the segment, or the selected
frame set for a single query.

\subsection{Adaptive visual evidence and test-time compute}

Most recent efficient video-LLM methods adapt \emph{which} visual evidence is
shown to the model, while a broader adaptive-compute literature studies
inference-time allocation more generally. Query-aware frame selection methods
score frames by their relevance to the current question, often combining
temporal coverage with visual-textual matching. Q-Frame selects frames and
adapts resolution according to the video content and query~\citep{zhang2025q};
GenS trains a generative frame sampler under capped frame budgets~\citep{yao2025generative};
and HiMu composes expert grounding signals for hierarchical frame
selection~\citep{ben2026himu}. In language-model serving and conditional
computation, FrugalGPT uses query-dependent cascades to reduce cost~\citep{chen2023frugalgpt},
RouteLLM routes between stronger and weaker LLMs as a cost-quality
tradeoff~\citep{ong2024routellm}, and Mixture-of-Depths allocates transformer
computation dynamically across tokens~\citep{raposo2024mixture}. 
Input-dependent mixture-of-experts routing has likewise been adopted beyond language modeling~\citep{xi2025multi}.
These works
share the premise that uniform computation is wasteful when inputs have
heterogeneous difficulty.

Our work connects these two perspectives in long-video QA. Rather than learning
a new retriever, selecting timestamps, or routing between different models,
FrameRouter reallocates a fixed dataset-level frame budget across questions
using frozen VLM inference configurations. Thus it is complementary to
within-question frame selection: those methods ask which evidence to keep,
whereas we ask how much visual evidence each question should receive under the
same average budget.

\section{Analysis: why route frame budgets?}
\label{sec:analysis}

\begin{figure}[h]

\centering
\subfigure[Mid-policy error recovery.]{
\begin{minipage}[t]{0.35\linewidth}
\centering
\resizebox{\linewidth}{!}{
\begin{tikzpicture}
\def\r{1.75}
\path[fill=blue!55, draw=white, line width=0.8pt]
  (0,0) -- (90:\r) arc[start angle=90, end angle=127.20, radius=\r] -- cycle;
\path[fill=teal!60, draw=white, line width=0.8pt]
  (0,0) -- (127.20:\r) arc[start angle=127.20, end angle=160.87, radius=\r] -- cycle;
\path[fill=orange!70, draw=white, line width=0.8pt]
  (0,0) -- (160.87:\r) arc[start angle=160.87, end angle=197.13, radius=\r] -- cycle;
\path[fill=gray!30, draw=white, line width=0.8pt]
  (0,0) -- (197.13:\r) arc[start angle=197.13, end angle=450, radius=\r] -- cycle;
\fill[white] (0,0) circle (0.82);
\node[font=\scriptsize, align=center] at (0,0)
  {Mid-policy\\Error};

\draw[black!45] (108.6:1.50) -- (-1.35,1.92);
\node[font=\scriptsize, align=center, anchor=south] at (-1.35,1.97)
  {Low\\$10.3\%$};

\draw[black!45] (144.0:1.50) -- (-2.02,0.98);
\node[font=\scriptsize, align=right, anchor=east] at (-2.10,0.98)
  {Both\\$9.4\%$};

\draw[black!45] (179.0:1.50) -- (-2.02,-0.48);
\node[font=\scriptsize, align=right, anchor=east] at (-2.10,-0.48)
  {High\\$10.1\%$};

\node[font=\scriptsize, align=center] at (315:1.30)
  {Unsolved\\$70.2\%$};

\end{tikzpicture}
}
\end{minipage}
\label{fig:analysis-error-recovery}
}
\hfill
\subfigure[Correction structure by question group.]{
\begin{minipage}[t]{0.6\linewidth}
\centering
\includegraphics[width=\linewidth]{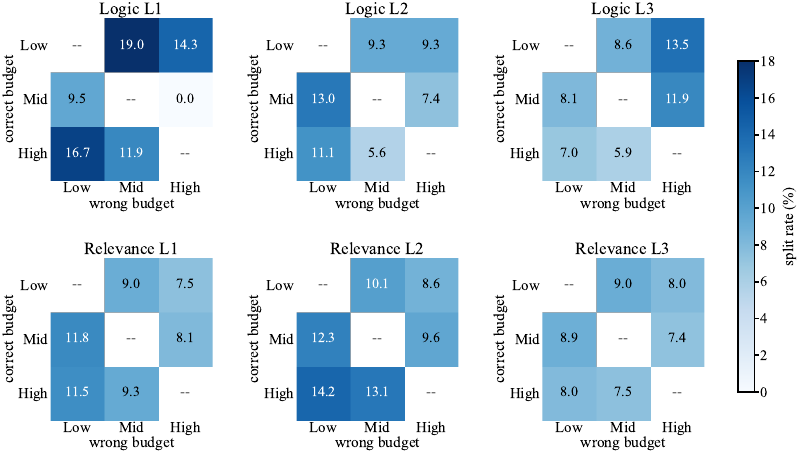}
\end{minipage}
\label{fig:analysis-correction-structure}
}
\caption{Question-level budget heterogeneity in the frozen Think
with-subtitles policy bank. Left: among errors made by the mid policy,
$29.8\%$ are answered correctly by at least one neighboring budget. Right:
pairwise correction matrices split by
\texttt{group\_type} and \texttt{level}; rows denote the policy that is correct
and columns denote the policy that is wrong.}
\label{fig:analysis-bank-diagnostics}
\end{figure}

We argue that uniform frame budgets are a poor unit of comparison on
Video-MME-v2, and that the right unit is a per-question budget assignment
chosen under a dataset-level mean-frame constraint. To make this argument
without conflating budget allocation with other design choices, we analyze a
\emph{frozen policy bank} of three full-video uniform sampling policies: low,
mid, and high budget, instantiated as 48, 64, and 96 frames. The policies use
the same video--language model, prompt, subtitles setting, and full-video
uniform sampler, differing only in the number of sampled frames. We use this
bank only as a diagnostic object: it lets us ask whether errors made by the
mid-budget reference are sometimes recoverable by neighboring budgets, and
whether such recovery has structure across question types. Any routed policy is
compared to the uniform reference under the same dataset-level mean frame
budget; the formal budget constraint is introduced in Section~\ref{sec:method}.
Figure~\ref{fig:analysis-bank-diagnostics} evaluates the question-level
headroom and structure of this bank in the main Think-with-subtitles setting on
the Video-MME-v2 full set.

\subsection{Mid-budget error recovery}
\label{sec:analysis:global}

Figure~\ref{fig:analysis-error-recovery} first asks whether the standard
mid-budget policy is merely making
irrecoverable errors, or whether some of its errors are artifacts of choosing
one uniform budget for every question. We use the mid policy as the
reference and inspect only its errors. This is an oracle diagnostic over the
frozen policy bank: it is not available to the deployed router, but it
measures the headroom available to a question-conditioned budget selector
restricted to the same bank.

The recoverable errors are split almost evenly between the two neighboring
budgets: low-only and high-only recovery occupy similar shares, with a smaller
overlap recovered by both neighboring policies. Thus the headroom is not only a
high-budget effect. Increasing the budget can also hurt: the high policy
introduces its own errors on cases that the mid policy answers correctly. The
frozen bank therefore does not define a nested error set in which more frames
are uniformly better. Instead, it defines an allocation problem: under the same
mid-budget mean, can we
promote the questions that need denser coverage while compressing questions
whose evidence is already sufficiently localized?

\subsection{Question-level disagreement}
\label{sec:analysis:disagree}

We next ask where this non-nested behavior concentrates. If disagreement were
uniformly scattered across the benchmark, a question-conditioned router would
have little structure to exploit. Instead, the correction patterns vary with
observable question groups and difficulty levels.

Figure~\ref{fig:analysis-correction-structure} shows that this disagreement
changes with both question group and difficulty level. We split the questions with official group and level labels by
$(\texttt{group\_type}, \texttt{level})$ and plot a small pairwise correction
matrix for each split: each entry is the percentage of questions
in that split for which the row budget is correct and the column budget is
wrong.

This pattern indicates that the three uniform policies do not form nested
error sets: they make systematically different mistakes at the question level.
We use this disagreement only as evidence that budget heterogeneity exists. A
deployable route 
must infer budget demand before answering, from the current question and
options alone.

\subsection{From diagnostics to design}
\label{sec:analysis:design}

These diagnostics suggest a narrow design space for a deployable router. To
isolate budget allocation from frame-selection effects, the route should choose
among the frozen full-video policies in $\Pi$ rather than introduce a new
sampler or temporal localizer. To avoid test-time leakage, the routing signal
should be computed from the current question prompt only, excluding labels,
videos, subtitles, sibling questions, other videos, model answers, and policy
outcomes. Because the route cannot observe which policy will be correct, the
signal should be an interpretable prior over evidence demand rather than a
direct correctness predictor. Finally, comparisons must enforce the same
dataset-level mean frame budget as the uniform reference; higher means are
separate operating points.

\section{FrameRouter}
\label{sec:method}

FrameRouter turns the question-level budget disagreement in
Section~\ref{sec:analysis} into a deployable routing method: it estimates
question-side evidence demand, routes each question to one frozen full-video
uniform policy, and enforces an exact mean-frame budget. It routes
\emph{frame budgets}, not frame locations, and never modifies the underlying
video--language model.

\subsection{Problem formulation}
\label{sec:method:problem}

Let $\mathcal{D}=\{(v_i, q_i, s_i, y_i)\}_{i=1}^{N}$ denote a video
question-answering dataset, where $v_i$ is a video, $q_i$ is the complete
question prompt including the answer options, $s_i$ is any auxiliary textual
context associated with the question (e.g., subtitles when provided by the
benchmark), and $y_i$ is the ground-truth answer. The auxiliary context $s_i$
may be used by the answer model, but it is not used as a routing signal. The
downstream task is multiple-choice video question answering: produce a
prediction $\hat{y}_i$ for each instance $i$. We treat the number of video
frames spent on instance $i$ as a problem-level variable $B_i\in\mathbb{N}$,
abstracting away the mechanism that produces those frames. A frame-budgeted
predictor is any procedure that, given $B_i$ frames of $v_i$ together with
$(q_i, s_i)$, returns an answer
\begin{equation}
\hat{y}_i \;=\; F\!\left(v_i,\,q_i,\,s_i;\,B_i\right).
\label{eq:budgeted-predictor}
\end{equation}
To make accuracy comparisons faithful to a fixed compute envelope, any
solution must spend, on average, exactly $B_{\mathrm{ref}}$ frames per
question. We denote the corresponding uniform reference policy by
$\pi_{\mathrm{ref}}$ and require
$\frac{1}{N}\sum_{i=1}^{N} B_i = B_{\mathrm{ref}}$. Operating points that
under- or over-spend this reference are different points on the cost--accuracy
plane and must be reported separately.
FrameRouter is a practical approximation to this fixed-budget assignment
problem: each parser call is question-only, and the solver uses only parser
scores and route quotas rather than labels or policy outcomes.

\subsection{Method overview}
\label{sec:method:framework}

FrameRouter instantiates the fixed-mean budget assignment problem with a
method-specific frozen policy bank
\begin{equation}
\Pi \;=\; \{\pi_{\mathrm{low}},\,\pi_{\mathrm{mid}},\,\pi_{\mathrm{high}}\},
\end{equation}
where each policy uniformly samples from the full video and uses the same
frozen video--language model and prompting protocol. The policies differ only
in their frame budgets, with
$B_{\mathrm{low}} < B_{\mathrm{mid}} < B_{\mathrm{high}}$. FrameRouter does
not modify the policy bank, the sampler, or the underlying model; it only
decides which bank member to use for each question.

The routing pipeline has two stages. First, an evidence-demand priority
parser assigns each instance two priorities: whether it is safe to compress to
the low-budget policy, and whether it is worth promoting to the high-budget
policy. Second, a quota-constrained allocator converts these priorities into
routes over $\Pi$ while preserving the same dataset-level mean frame budget as
the uniform reference. At inference time, FrameRouter executes only the
selected policy for each question; the other bank policies are not run, so the
method performs budget routing rather than late selection over candidate
answers.

\subsection{Evidence-demand priority parser}
\label{sec:method:parser}

The analysis in Section~\ref{sec:analysis} shows that budget effects vary
across questions rather than following a simple high-budget-fixes-low-budget
pattern. Because group labels, videos, subtitles, and policy outcomes are not
available to a deployed router, FrameRouter estimates evidence demand from the
question prompt itself. To keep this signal interpretable and auditable, it
does not directly predict which policy will answer correctly. Instead, a frozen
text-only parser assigns two priorities to the current question:
\begin{equation}
g_\phi(q_i) \;=\; \left(p_i^{\mathrm{low}},\,p_i^{\mathrm{high}}\right),
\label{eq:priority-parser}
\end{equation}
where $p_i^{\mathrm{low}},p_i^{\mathrm{high}}\in[0,1]$ are direction-specific
scores rather than a normalized distribution. The low-priority score estimates
how safe it is to compress instance $i$ to the low-budget policy, for example
when the answer appears anchored to local, static, single-moment, or explicit
textual evidence. The high-priority score estimates how useful additional
coverage may be, for example for counting, ordering, change tracking, broad
temporal evidence, or fine visual discrimination across moments. For auditing,
the same parser also returns auxiliary fields such as confidence,
evidence-anchor category, optional route suggestion, and a short rationale;
these fields are not consumed by the allocator.

The parser prompt and underlying LLM checkpoint, denoted by $\phi$, are fixed
before evaluation and are never updated using benchmark labels, policy
outcomes, or category-level performance. The parser uses deterministic decoding
and sees only $q_i$, including the multiple-choice options that are part of the
task input. It does not see the video, subtitles, sibling questions, other
videos, the correct answer label, or any candidate answers generated by the
policy bank. Thus the priority pair is a question-only prior over evidence
demand, not a temporal localization claim: FrameRouter never predicts
timestamps, and all policies in $\Pi$ remain full-video uniform samplers.

\subsection{Quota-constrained allocation}
\label{sec:method:allocation}

Given the priority scores produced by $g_\phi$ in Eq.~\eqref{eq:priority-parser},
FrameRouter must turn them into routes $\{r_i\}_{i=1}^{N}$ that satisfy the
dataset-level mean-budget constraint
$\frac{1}{N}\sum_{i=1}^{N} B_{r_i}=B_{\mathrm{ref}}$ exactly. We do this through a
quota-constrained allocator that decouples \emph{how many} questions are
sent to each policy from \emph{which} questions are sent there. The first
choice is fixed by an operating-point quota under the budget constraint; the
second is determined by two priority rankings.

\paragraph{Quota selection.}
Let $n_{\mathrm{low}}, n_{\mathrm{mid}}, n_{\mathrm{high}}\in\mathbb{Z}_{\ge 0}$
denote the number of instances assigned to each policy in $\Pi$. For a chosen
operating point, FrameRouter fixes a feasible quota triple with
$n_{\mathrm{low}} + n_{\mathrm{mid}} + n_{\mathrm{high}} = N$ and
$(B_{\mathrm{low}}n_{\mathrm{low}} +
B_{\mathrm{mid}}n_{\mathrm{mid}} +
B_{\mathrm{high}}n_{\mathrm{high}})/N = B_{\mathrm{ref}}$. This encodes
the mean-budget constraint exactly at the dataset level rather than only in
expectation. Specializing to the common setting
$B_{\mathrm{ref}}=B_{\mathrm{mid}}$, this condition is equivalently
\begin{equation}
(B_{\mathrm{mid}}-B_{\mathrm{low}}) n_{\mathrm{low}}
=
(B_{\mathrm{high}}-B_{\mathrm{mid}}) n_{\mathrm{high}},
\label{eq:quota-balance}
\end{equation}
with $n_{\mathrm{mid}}=N-n_{\mathrm{low}}-n_{\mathrm{high}}\ge 0$. Thus
feasible quotas form an integer family that trades low-budget assignments
against high-budget assignments while preserving the reference mean. The
quota triple is fixed before evaluation and does not use labels, policy
correctness, or category-level performance. It is selected as a pre-specified
operating point under the budget equation rather than by test-set accuracy.
Different feasible triples correspond to different routing operating points.

\paragraph{Deterministic assignment.}
Given $(n_{\mathrm{low}}, n_{\mathrm{mid}}, n_{\mathrm{high}})$, we assign
routes with a high-first, low-second rule. Let
$\operatorname{argTopK}_{i\in S}^{m} \tau_i$ denote the index set of the $m$
largest scores in $S$, breaking ties deterministically with a fixed hash of
the dataset index. FrameRouter first selects
\begin{equation}
H = \operatorname{argTopK}_{i\in\{1,\ldots,N\}}^{n_{\mathrm{high}}} p_i^{\mathrm{high}},
\end{equation}
and assigns $r_i=\mathrm{high}$ for $i\in H$. It then selects the low-budget
set from the remaining instances,
\begin{equation}
L = \operatorname{argTopK}_{i\in\{1,\ldots,N\}\setminus H}^{n_{\mathrm{low}}} p_i^{\mathrm{low}},
\end{equation}
and assigns $r_i=\mathrm{low}$ for $i\in L$. All remaining instances are routed
to the mid-budget policy. The high-first order is intentional: if an ambiguous
question receives high scores for both the low- and high-policy priorities,
the allocator treats the high-policy signal as the stronger constraint: wrongly
compressing a high-demand question can remove needed evidence, whereas wrongly
expanding a low-demand question mainly spends redundant frames. The construction
guarantees, by the quota constraint, that
$\frac{1}{N}\sum_{i=1}^{N} B_{r_i} = B_{\mathrm{ref}}$ holds exactly, so an
accuracy gap against the uniform reference $\pi_{\mathrm{ref}}$ is not explained
by a change in average compute, but by how that compute is distributed across
questions. Inference then executes only the selected policy $\pi_{r_i}$ for
instance $i$; the other
policies in $\Pi$ are not run, so there is no late selection over multiple
candidate answers. The parser adds one short text-only LLM call per question,
which is separate from the video-frame budget.

\section{Experiments}
\label{sec:experiments}

We evaluate FrameRouter on Video-MME-v2 and three additional video understanding benchmarks, always matching the mean frame budget of the corresponding uniform mid-budget baseline.

\begin{table}[t]
\centering
\caption{Main results on Video-MME-v2 under the same 64-frame mean-budget protocol in \%. We report the group-level metrics and average accuracy with and without subtitles.}
\label{tab:videommev2_main_results}
\definecolor{SubCol}{RGB}{232,244,253}
\newcommand{\subcol}[1]{\cellcolor{SubCol}{#1}}
\setlength{\tabcolsep}{1.15pt}
\renewcommand{\arraystretch}{1.26}
\normalsize
\makebox[\textwidth][c]{%
\resizebox{\textwidth}{!}{%
\begin{tabular}{@{}l c *{7}{cc}@{}}
\toprule
\multirow{2}{*}{\raisebox{-0.55ex}{\textbf{Model}}} & \multirow{2}{*}{\raisebox{-0.55ex}{\textbf{Frames}}}
& \multicolumn{2}{c}{\textbf{Non-Lin}}
& \multicolumn{2}{c}{\textbf{Level 1}}
& \multicolumn{2}{c}{\textbf{Level 2}}
& \multicolumn{2}{c}{\textbf{Level 3}}
& \multicolumn{2}{c}{\textbf{Consistency}}
& \multicolumn{2}{c}{\textbf{Coherence}}
& \multicolumn{2}{c}{\textbf{Avg. Acc.}} \\
\cmidrule(lr){3-4}
\cmidrule(lr){5-6}
\cmidrule(lr){7-8}
\cmidrule(lr){9-10}
\cmidrule(lr){11-12}
\cmidrule(lr){13-14}
\cmidrule(l){15-16}
& & \subcol{\textbf{w. sub}} & \textbf{wo sub}
& \subcol{\textbf{w. sub}} & \textbf{wo sub}
& \subcol{\textbf{w. sub}} & \textbf{wo sub}
& \subcol{\textbf{w. sub}} & \textbf{wo sub}
& \subcol{\textbf{w. sub}} & \textbf{wo sub}
& \subcol{\textbf{w. sub}} & \textbf{wo sub}
& \subcol{\textbf{w. sub}} & \textbf{wo sub} \\
\midrule
\multicolumn{16}{@{}c@{}}{\textbf{Commercialization Models}} \\
\midrule
Gemini-3-Pro & 1fps & \subcol{49.4} & 38.2 & \subcol{64.0} & 43.2 & \subcol{50.0} & 45.4 & \subcol{40.6} & 30.2 & \subcol{50.8} & 39.7 & \subcol{47.0} & 35.4 & \subcol{66.1} & 56.8 \\
GPT-5 & 50 & \subcol{37.0} & 26.4 & \subcol{44.5} & 32.2 & \subcol{39.1} & 28.6 & \subcol{31.1} & 21.3 & \subcol{37.4} & 28.1 & \subcol{36.3} & 23.2 & \subcol{55.6} & 44.7 \\
Kimi-K2.5 & 64 & \subcol{36.1} & 27.3 & \subcol{44.3} & 30.5 & \subcol{40.0} & 32.8 & \subcol{28.5} & 21.6 & \subcol{36.0} & 28.7 & \subcol{36.2} & 24.9 & \subcol{54.4} & 46.0 \\
\midrule
\multicolumn{16}{@{}c@{}}{\textbf{Open-Source Models (Thinking)}} \\
\midrule
MiMo-VL-7B-RL & 64 & \subcol{19.4} & 11.9 & \subcol{24.9} & 16.1 & \subcol{18.8} & 12.5 & \subcol{16.6} & 9.0 & \subcol{18.9} & 12.5 & \subcol{20.3} & 10.8 & \subcol{36.0} & 27.0 \\
GLM4.5V & 64 & \subcol{21.8} & 13.3 & \subcol{26.2} & 14.9 & \subcol{21.7} & 14.7 & \subcol{19.4} & 11.4 & \subcol{20.5} & 14.7 & \subcol{24.3} & 10.6 & \subcol{39.2} & 29.6 \\
Qwen3-VL-235B-A22B & 64 & \subcol{26.3} & 15.9 & \subcol{32.0} & 19.1 & \subcol{27.5} & 16.8 & \subcol{22.3} & 13.4 & \subcol{26.0} & 17.2 & \subcol{27.0} & 13.4 & \subcol{44.9} & 33.3 \\
\midrule
Qwen3.5-397B-A17B & 64 & \subcol{30.6} & 20.7 & \subcol{34.8} & 20.7 & \subcol{30.4} & 23.1 & \subcol{28.4} & 19.0 & \subcol{29.5} & 20.9 & \subcol{32.7} & 20.4 & \subcol{48.9} & 39.3 \\
\textbf{FrameRouter} & \textbf{64} & \subcol{\textbf{34.9}} & \textbf{24.6} & \subcol{\textbf{38.8}} & \textbf{26.1} & \subcol{\textbf{38.1}} & \textbf{29.6} & \subcol{\textbf{30.3}} & \textbf{20.0} & \subcol{\textbf{33.2}} & \textbf{25.6} & \subcol{\textbf{37.9}} & \textbf{22.6} & \subcol{\textbf{53.3}} & \textbf{42.9} \\
\midrule
\multicolumn{16}{@{}c@{}}{\textbf{Open-Source Models (Instruct)}} \\
\midrule
MiMo-VL-7B-SFT & 64 & \subcol{14.7} & 10.7 & \subcol{19.8} & 14.3 & \subcol{13.2} & 9.6 & \subcol{12.8} & 9.3 & \subcol{14.1} & 11.2 & \subcol{15.8} & 9.7 & \subcol{31.2} & 24.6 \\
LLaVA-Video-72B & 64 & \subcol{17.2} & 11.3 & \subcol{21.8} & 14.8 & \subcol{14.8} & 11.1 & \subcol{16.3} & 9.5 & \subcol{17.1} & 12.3 & \subcol{17.5} & 9.5 & \subcol{34.4} & 27.3 \\
Qwen2.5-VL-72B & 64 & \subcol{22.1} & 13.7 & \subcol{25.8} & 15.3 & \subcol{21.2} & 13.7 & \subcol{20.6} & 12.6 & \subcol{20.9} & 14.1 & \subcol{24.3} & 12.9 & \subcol{38.9} & 30.3 \\
InternVL3.5-241B-A28B & 64 & \subcol{23.1} & 15.8 & \subcol{28.2} & 18.1 & \subcol{23.7} & 16.3 & \subcol{19.6} & 14.1 & \subcol{23.4} & 17.2 & \subcol{22.5} & 13.2 & \subcol{41.4} & 32.9 \\
Qwen3-VL-235B-A22B & 64 & \subcol{25.0} & 16.5 & \subcol{30.7} & 20.6 & \subcol{25.2} & 16.8 & \subcol{21.6} & 13.9 & \subcol{25.0} & 17.7 & \subcol{25.0} & 14.2 & \subcol{43.3} & 33.8 \\
\midrule
Qwen3.5-397B-A17B & 64 & \subcol{24.5} & 16.9 & \subcol{27.3} & 16.8 & \subcol{25.9} & 18.1 & \subcol{21.9} & 16.1 & \subcol{23.9} & 18.2 & \subcol{25.7} & 14.6 & \subcol{42.1} & 35.0 \\
\textbf{FrameRouter} & \textbf{64} & \subcol{\textbf{25.7}} & \textbf{18.5} & \subcol{\textbf{30.6}} & \textbf{21.3} & \subcol{\textbf{27.2}} & \textbf{19.2} & \subcol{\textbf{21.8}} & \textbf{16.3} & \subcol{\textbf{26.2}} & \textbf{19.6} & \subcol{\textbf{24.9}} & \textbf{16.4} & \subcol{\textbf{44.8}} & \textbf{36.8} \\
\bottomrule
\end{tabular}%
}}

\end{table}

\subsection{Setup}
\label{sec:experiments:setup}

\paragraph{Benchmarks and metrics.}
We evaluate on four video understanding benchmarks: Video-MME-v2~\citep{fu2026video}, LongVideoBench~\citep{wu2024longvideobench}, MVBench~\citep{li2024mvbench}, and VideoMMMU~\citep{hu2025video}. Video-MME-v2 provides the most detailed protocol for our main analysis: in addition to average accuracy, it reports official group-level scores for non-linear reasoning, three difficulty levels, consistency, and coherence, and we evaluate both subtitle and no-subtitle settings. LongVideoBench is also evaluated in both subtitle settings, while MVBench and VideoMMMU are evaluated with their official accuracy metrics. Unless otherwise noted, all reported numbers follow the corresponding benchmark evaluation protocol.

\paragraph{Baselines.}
Our controlled baseline is uniform mid-budget inference with the same Qwen3.5-397B-A17B backbone~\citep{qwen2026qwen35}. We evaluate both \emph{Instruct} and \emph{Think} regimes, and constrain FrameRouter to the same average visual frame budget using the frozen 48/64/96 full-video policy bank. The parser is a deterministic text-only call to the same model used by the answer-stage VLM, which receives only question/options, processes no video frames, and is capped at 160 output tokens, so its overhead is negligible compared with the heavy multimodal answer call over 48/64/96 frames and up to 1024 generated tokens. For context, we also include frontier commercial systems such as Gemini-3-Pro~\citep{google2025gemini3pro}, GPT-5~\citep{openai2025gpt5}, and Kimi-K2.5~\citep{kimi2026k25}, together with open-source video LMMs such as Qwen3-VL~\citep{bai2025qwen3vl}, LLaVA-Video~\citep{zhang2024llavavideo}, InternVL3.5~\citep{wang2025internvl35}, and Qwen2.5-VL~\citep{bai2025qwen25vl}.

\subsection{Main results}
\label{sec:experiments:videommev2}
\begin{table}[t]
\centering
\caption{Comparison across long-video and video understanding benchmarks in \%. Missing or unreported entries are denoted by ``-''.}
\label{tab:cross_benchmark_results}
\setlength{\tabcolsep}{3.2pt}
\renewcommand{\arraystretch}{1.14}
\footnotesize
\begin{tabular}{@{}l *{4}{cc}@{}}
\toprule
\multirow{2}{*}{\raisebox{-0.55ex}{\textbf{Model}}}
& \multicolumn{2}{c}{\textbf{MVBench}} 
& \multicolumn{2}{c}{\textbf{VideoMMMU}} 
& \multicolumn{2}{c}{\begin{tabular}{@{}c@{}}\textbf{L.V.Bench}\\[-0.25ex]\textbf{(w. sub)}\end{tabular}} 
& \multicolumn{2}{c}{\begin{tabular}{@{}c@{}}\textbf{L.V.Bench}\\[-0.25ex]\textbf{(wo sub)}\end{tabular}} \\
\cmidrule(lr){2-3}
\cmidrule(lr){4-5}
\cmidrule(lr){6-7}
\cmidrule(l){8-9}
& \textbf{Frames} & \textbf{Acc.}
& \textbf{Frames} & \textbf{Acc.}
& \textbf{Frames} & \textbf{Acc.}
& \textbf{Frames} & \textbf{Acc.} \\
\midrule
\multicolumn{9}{@{}c@{}}{\textbf{Commercialization Models}} \\
\midrule
Gemini-3-Pro & - & 74.1 & - & 87.6 & - & 77.7 & - & - \\
GPT-5 & 50 & 74.1 & 50 & 84.6 & 50 & 72.6 & - & - \\
Kimi-K2.5 & - & 73.5 & - & 86.6 & - & 79.8 & - & - \\
\midrule
\multicolumn{9}{@{}c@{}}{\textbf{Open-Source Models (Thinking)}} \\
\midrule
GLM4.5V & - & 73.0 & - & 72.4 & 64 & 68.8 & - & - \\
MiMo-VL-7B-RL & - & - & 256 & 53.1 & - & - & - & - \\
Qwen3-VL-235B-A22B & - & 75.2 & - & 80.0 & - & 65.6 & - & - \\
\midrule
Qwen3.5-397B-A17B & 64 & 74.2 & 64 & 84.7 & 64 & 70.4 & 64 & 68.1 \\
\textbf{FrameRouter} & \textbf{64} & \textbf{75.4} & \textbf{64} & \textbf{86.0} & \textbf{64} & \textbf{73.1} & \textbf{64} & \textbf{71.0} \\
\midrule
\multicolumn{9}{@{}c@{}}{\textbf{Open-Source Models (Instruct)}} \\
\midrule
LLaVA-Video-72B & - & 64.1 & - & 49.7 & - & - & 64 & 61.9 \\
Qwen2.5-VL-72B & 16 & 70.4 & - & 60.2 & 64 & 60.7 & - & - \\
InternVL3.5-241B-A28B & 16 & 76.5 & - & - & 64 & 67.1 & - & - \\
\midrule
Qwen3.5-397B-A17B & 64 & 74.0 & 64 & 73.3 & 64 & 69.7 & 64 & 66.8 \\
\textbf{FrameRouter} & \textbf{64} & \textbf{74.2} & \textbf{64} & \textbf{75.1} & \textbf{64} & \textbf{72.1} & \textbf{64} & \textbf{69.7} \\
\bottomrule
\end{tabular}%

\end{table}

\paragraph{Video-MME-v2.}
Table~\ref{tab:videommev2_main_results} reports Video-MME-v2 results at the
same 64-frame mean budget. The strongest controlled gain appears in the Think
regime with subtitles: relative to the uniform Qwen3.5-397B-A17B row,
FrameRouter improves average accuracy by $4.4\%$ and raises every
group-level metric. The largest gains concentrate on the harder diagnostic
axes: Level~2 improves by $7.7\%$, Coherence by $5.2\%$, and Non-Lin
by $4.3\%$. This routed Think setting is also the strongest open-source Think
row in the table while using the same 64-frame mean budget. The no-subtitle and
Instruct settings show the same overall direction, indicating that the benefit
is not confined to a single input or decoding configuration.

\paragraph{Additional benchmarks.}
Table~\ref{tab:cross_benchmark_results} gives the corresponding results on
MVBench, VideoMMMU, and LongVideoBench. In the Think block, FrameRouter
improves over the same Qwen3.5-397B-A17B uniform row on every reported
benchmark, with gains ranging from roughly $1\%$ on shorter or more
saturated settings to nearly $3\%$ on LongVideoBench. The routed
LongVideoBench Think result with subtitles also exceeds the listed GPT-5
reference in Table~\ref{tab:cross_benchmark_results}. The Instruct block shows
a similar positive pattern, with the clearest gains appearing on VideoMMMU and
LongVideoBench. Across
Table~\ref{tab:videommev2_main_results} and
Table~\ref{tab:cross_benchmark_results}, the most consistent message is that
fixed-budget routing transfers across benchmarks and is especially effective
on long-video and reasoning-heavy settings.

\subsection{Ablation study}
\label{sec:experiments:ablation}

We ablate the two ingredients most specific to FrameRouter: direction-specific
priority scores and strict-budget quota allocation. All variants use the same
Think-with-subtitles setting on Video-MME-v2 and keep the mean frame budget
matched to the uniform baseline.

\paragraph{Score channels.}
Table~\ref{tab:ablation_score_channels} shows that both priority channels
contribute useful routing signal. Using the full direction-specific parser
improves average accuracy by $4.4\%$ over uniform inference, with larger gains
on the diagnostic metrics most tied to long-video reasoning: Coherence improves
by $5.2\%$ and Non-Lin by $4.3\%$. The single-channel variants also improve
substantially, indicating that each score captures a meaningful part of
evidence demand; combining them yields the most consistent profile across the
reported metrics.

\begin{figure}[t]
\centering
\subfigure[Quota sensitivity.]{
    \begin{minipage}[t]{0.48\linewidth}
    \centering
    \begin{tikzpicture}
\begin{axis}[
    name=accaxis,
    width=\linewidth,
    height=0.71\linewidth,
    xmin=180,
    xmax=920,
    ymin=52.5,
    ymax=53.45,
    xlabel={Quota $q$},
    ylabel={Avg.\ Acc. (\%)},
    xtick={200,300,400,500,600,700,800,900},
    tick align=inside,
    label style={font=\scriptsize},
    tick label style={font=\tiny},
    legend style={at={(0.02,0.98)}, anchor=north west, draw=none, fill=none, font=\tiny},
    legend cell align=left,
    y tick label style={
        font=\tiny,
        /pgf/number format/.cd,
        fixed,
        fixed zerofill,
        precision=1,
    },
    grid=major,
    grid style={dashed, gray!30},
]
\addplot[mark=*, color=blue, thick] coordinates{
    (200,52.63) (250,52.74) (300,52.69) (350,52.75)
    (400,52.77) (450,53.08) (500,53.18) (550,53.34)
    (600,53.14) (650,53.27) (700,53.05) (750,53.24)
    (800,53.22) (850,53.23) (900,52.87)
};
\addlegendentry{Avg.\ Acc.}
\addlegendimage{mark=square*, color=orange, thick}
\addlegendentry{Non-Lin}
\end{axis}
\begin{axis}[
    width=\linewidth,
    height=0.71\linewidth,
    xmin=180,
    xmax=920,
    ymin=33.7,
    ymax=35.0,
    axis x line=none,
    axis y line*=right,
    ylabel={Non-Lin (\%)},
    xtick=\empty,
    tick align=inside,
    label style={font=\scriptsize},
    tick label style={font=\tiny},
    y tick label style={
        font=\tiny,
        /pgf/number format/.cd,
        fixed,
        fixed zerofill,
        precision=1,
    },
]
\addplot[mark=square*, color=orange, thick] coordinates{
    (200,34.00) (250,34.00) (300,34.02) (350,33.93)
    (400,33.82) (450,34.09) (500,34.44) (550,34.86)
    (600,34.58) (650,34.66) (700,34.40) (750,34.50)
    (800,34.51) (850,34.56) (900,34.38)
};
\end{axis}
\end{tikzpicture}
    \label{fig:quota-sensitivity}
    \end{minipage}

}
\hfill
\subfigure[Route distribution by group$\times$level.]{
    \begin{minipage}[t]{0.48\linewidth}
    \centering
    \includegraphics[width=\linewidth]{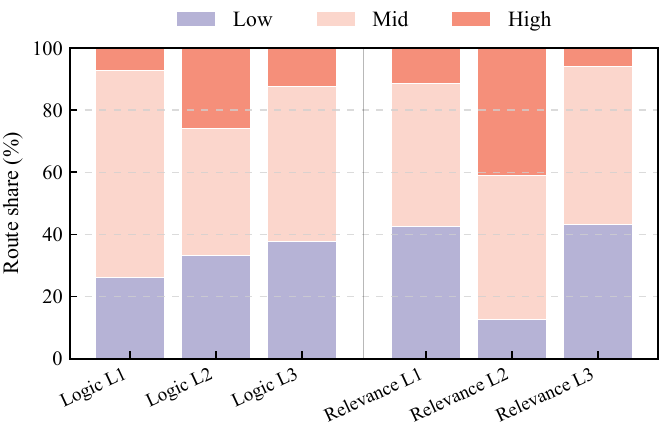}
    \label{fig:router-group-level-distribution}
    \end{minipage}
}
\caption{Allocator diagnostics on Video-MME-v2 Think with subtitles. Left:
quota sensitivity under the matched mean budget. Right: realized route shares
within each official group$\times$level split.}
\label{fig:allocator-diagnostics}
\end{figure}

\paragraph{Quota sensitivity.}
Figure~\ref{fig:quota-sensitivity} sweeps the high-route quota while preserving
the same mean frame budget. The selected operating point lies inside a broad
high-performing region rather than at an isolated spike, and nearby quotas
change the reported metrics by only a few tenths of a percent. The two curves
also show that the gain is not explained by simply assigning more questions to
the high-budget policy; under a fixed mean budget, useful allocation requires a
balanced redistribution between compressed and promoted questions.

\begin{table}[t]
\centering
\caption{Score-channel ablation on Video-MME-v2 with the Think backbone and subtitles. All routed variants use the same route quota and mean frame budget; in single-channel variants, the enabled score supplies the ranking signal for both quota-constrained selections.}
\label{tab:ablation_score_channels}
\setlength{\tabcolsep}{4.2pt}
\renewcommand{\arraystretch}{1.12}
\small
\begin{tabular}{@{}lcc cccc@{}}
\toprule
Method & $p^{\mathrm{low}}$ & $p^{\mathrm{high}}$ & Non-Lin & Consistency & Coherence & Avg. Acc. \\
\midrule
Baseline & $\times$ & $\times$ & 30.60 & 29.50 & 32.70 & 48.90 \\
High-only & $\times$ & $\checkmark$ & 33.94 & 33.10 & 35.47 & 53.03 \\
Low-only & $\checkmark$ & $\times$ & 34.64 & 32.86 & 37.91 & 52.66 \\
\textbf{Ours} & $\checkmark$ & $\checkmark$ & \textbf{34.86} & \textbf{33.20} & \textbf{37.93} & \textbf{53.34} \\
\bottomrule
\end{tabular}
\end{table}

\subsection{Router behavior analysis}
\label{sec:experiments:router_behavior}

Route assignments in FrameRouter are determined before answer generation from
the current question alone, without access to the video, subtitles, labels, or
candidate-policy outputs. We therefore use route behavior only as a post-hoc
diagnostic of whether the question-only parser induces structured budget
allocation under the fixed mean budget.

Figure~\ref{fig:router-group-level-distribution} shows a structured allocation
pattern across the official group$\times$level splits. The mid route remains
the dominant fallback, while the low- and high-budget shares vary across
question categories rather than following a uniform mixture. This supports the
central premise of FrameRouter: even without observing video content or
outcomes, question prompts contain enough evidence-demand signal to redistribute
a fixed visual budget in a category-sensitive way.

\section{Conclusion}

This work studies whether a uniform per-question frame budget is the best use
of a fixed average visual budget for long-video question answering. Our
diagnostics on Video-MME-v2 show that nearby frame budgets help different
question subsets, motivating fixed-mean frame-budget routing as a controlled
test-time setting. FrameRouter instantiates this setting with a training-free
question-only evidence-demand parser and a quota-constrained allocator over a
frozen bank of full-video uniform policies. Across Video-MME-v2,
LongVideoBench, MVBench, and VideoMMMU, the results show that matched-budget
gains can be obtained by reallocating visual evidence across heterogeneous
questions rather than increasing the average frame count. Budget routing is
therefore a simple inference axis complementary to frame selection, token
compression, and model scaling.

\begin{ack}
\textbf{Funding.} This work received no external funding.

\textbf{Competing interests.} The authors declare no competing interests.
\end{ack}

\bibliographystyle{plainnat}
\bibliography{references}

\newpage
\appendix
\section{Technical appendix}
\label{sec:appendix-diagnostics}

\subsection{Additional policy-bank corrections}
\label{sec:appendix-instruct-corrections}

Section~\ref{sec:analysis:disagree} analyzes the Think setting and
shows that the low-, mid-, and high-budget policies do not form nested error
sets. Here we repeat the same diagnostic for the Instruct regime to verify
that budget complementarity is not specific to chain-of-thought style
inference. For each official group$\times$level split, we compute pairwise
correction matrices among the frozen full-video uniform policies. Rows denote
the policy that answers a question correctly, columns denote the policy that
misses it, and each mini-matrix is normalized within its split. Off-diagonal
mass therefore indicates recoverable disagreement: questions that one budget
handles while another budget fails.

This diagnostic is intentionally separated from the routed results. It does
not use the parser, the allocator, or any learned route prediction; it only
examines the frozen policy bank under different frame budgets. The Instruct setting shows
that budget disagreement remains structured even without explicit reasoning
traces, suggesting that frame-budget complementarity is a property of the
question--evidence interaction rather than an artifact of a particular decoding
style.

\begin{figure}[h]
\centering
\includegraphics[width=0.95\linewidth]{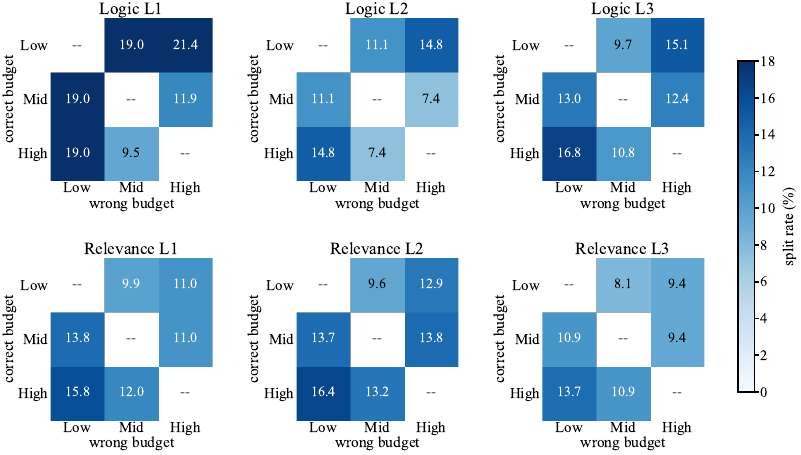}
\caption{Pairwise correction matrices split by \texttt{group\_type} and
\texttt{level} in the Instruct setting. We show the labeled L1--L3 splits for
logic and relevance questions; Video-MME-v2 logic questions with empty level
fields are excluded from this diagnostic plot. Each mini-matrix is normalized
within its split; rows denote the policy that is correct and columns denote
the policy that is wrong. The off-diagonal mass varies across question groups
and difficulty levels, supporting question-conditioned budget routing rather
than a single monotone frame-scaling rule.}
\label{fig:category-corrections-instruct}
\end{figure}

Figure~\ref{fig:category-corrections-instruct} repeats the group$\times$level
pairwise correction analysis for the Instruct regime. The off-diagonal entries
remain visibly non-uniform across groups and difficulty levels, meaning that
the low-, mid-, and high-budget policies recover different subsets of questions
rather than forming a simple nested ordering. This mirrors the Think-setting
diagnostic in the main text and supports the same design choice: FrameRouter
should route among frozen full-video budget policies instead of assuming that a
single uniform budget is uniformly preferable for all question types.

\subsection{Parser evidence-anchor profiles}
\label{sec:appendix-anchor-profiles}

Section~\ref{sec:experiments:router_behavior} reports how the final routes
distribute across official group$\times$level splits. The diagnostic below
looks at the same routed questions from the parser side. During routing, the
question-only parser emits an evidence-anchor category together with its
low- and high-budget priorities. These categories are not used as ground-truth
annotations and are not meant to localize evidence in the video; they are
self-reported parser fields that help interpret why different questions are
assigned to different budget routes. We aggregate these anchor categories
within each final route to check whether the route assignments correspond to
coherent evidence-demand profiles.

This analysis should be read as an interpretability check rather than as an
additional source of supervision. The allocator consumes only the two priority
scores described in Section~\ref{sec:method:parser}; the evidence-anchor field
is logged for auditing and is not used to choose the route. A meaningful
anchor--route association is therefore useful because it shows that the parser
priorities are not arbitrary scalar scores: they are accompanied by stable,
human-readable evidence profiles that align with the intended roles of the low,
mid, and high routes.

\begin{figure}[h]
\centering
\includegraphics[width=0.6\linewidth]{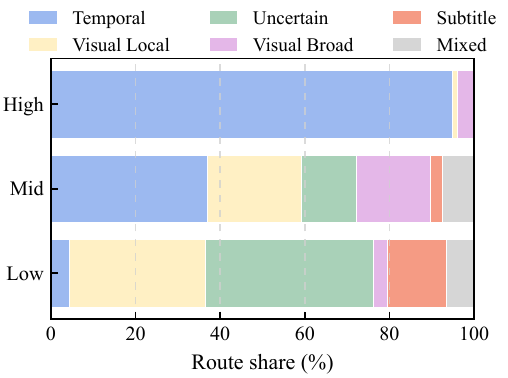}
\caption{Parser evidence-anchor profiles within each route on Video-MME-v2 Think with subtitles. Routes are decided from the current question only, prior to answer generation. Evidence-anchor labels are parser outputs and are used only to interpret routing behavior.}
\label{fig:router-anchor-distribution-appendix}
\end{figure}

Figure~\ref{fig:router-anchor-distribution-appendix} gives a parser-side view
of the realized routes. The high route is strongly enriched for temporal
evidence-anchor outputs, while the low route contains more uncertain,
local-visual, and subtitle-oriented profiles. The mid route remains more mixed,
consistent with its role as the fallback route when neither compression nor
promotion is strongly preferred. These anchors are not human annotations and do
not imply temporal localization, but they help verify that the question-only
parser induces interpretable route structure rather than an arbitrary budget
mixture.

\subsection{Implementation details}
\label{sec:appendix-implementation}

This section summarizes implementation details of our method. The default configuration matches the main
Video-MME-v2 Think-with-subtitles setting unless otherwise specified.

\paragraph{Hyperparameters.}
FrameRouter uses a three-policy full-video uniform bank with low, mid, and
high budgets set to 48, 64, and 96 frames, respectively. For the 48/64/96 bank this gives twice as many
low routes as high routes. Parser and answer calls use temperature $0.0$ by default. The parser response
length is 160 tokens, while the answer response length is 1024 tokens.

\paragraph{Deterministic allocation and parsing.}
The parser returns \texttt{priority\_48}, \texttt{priority\_96},
\texttt{confidence}, \texttt{evidence\_anchor}, and a short rationale. Scores
are clamped to $[0,1]$, and malformed or missing scores default to conservative
mid-range values. The allocator first selects the high route by descending
\texttt{priority\_96}, then selects the low route from the remaining questions
by descending \texttt{priority\_48}; all remaining questions use the mid route.
Ties are broken by a deterministic hash of the dataset index, and the code
raises an error if the allocated mean frame count differs from the mid-budget
target. The answer stage then runs only the selected route for each question,
using uniform sampling and the baseline Video-MME-v2 prompt layout inherited
from the evaluation script.

\paragraph{Evidence-demand priority parser prompt.}
The released parser prompt is presented below. The prompt explicitly states
the information boundary of the router, the fixed-mean budget setting, the
semantics of the two priority scores, and the required JSON output format.

\begin{center}
\begin{minipage}{0.99\linewidth}
\begin{Verbatim}[frame=single,framesep=2mm,fontsize=\small]
You are a text-only priority router for long-video multiple-choice QA.

You cannot see the video and you do not know the answer. Use only the current
question and answer options. Do not answer the question. Do not use labels,
sibling questions, other videos, or dataset statistics.

A strict solver will keep the average budget fixed at 64 frames:
- a small top group by priority_96 will use 96 frames;
- twice as many questions by priority_48 will use 48 frames;
- all remaining questions will use the default 64 frames.

Your job is to output the two final ranking priorities directly. Do not output a
weighted formula; do not choose the number of questions assigned to each budget.

priority_48 should be high only when reducing 64f to 48f is likely safe:
- the decisive evidence is local, static, repeated, or easy to locate;
- 64f looks more than sufficient from the question/options;
- the question is anchored by a clear subtitle/dialogue/text/object cue;
- no broad temporal coverage, counting, order tracking, or subtle multi-scene
  comparison is needed.

priority_96 should be high only when extra frames are likely useful:
- subtitle/dialogue/text anchoring is weak or absent;
- the decisive visual evidence is hard to localize from text alone;
- broad temporal coverage, counting, event order, changes, or repeated actions
  are central;
- the options require fine visual discrimination across moments.

Important calibration:
- Most questions should remain 64f, so both priorities should often be modest.
- Do not make priority_96 high merely because the question seems hard; extra
  frames only help when the missing evidence is temporal/visual coverage.
- Do not make priority_48 high for OCR, tiny text, rare brief details, or fine
  visual distinctions even if the evidence is local.
- If unsure, set both priorities near 0.40-0.55 and confidence below 0.60.

Return compact JSON only:
{
"policy": "DEFAULT_64F",
"priority_48": 0.00,
"priority_96": 0.00,
"confidence": 0.00,
"evidence_anchor": "subtitle|visual_local|visual_broad|temporal|mixed|uncertain",
"rationale": "<=10 words"
}

Question:
{question}

Options:
{options}
\end{Verbatim}
\end{minipage}
\end{center}

\paragraph{Answer-stage prompts.}
The answer stage uses the baseline Video-MME-v2 prompt layout from the
official evaluation script.

\subsection{Code release}
\label{sec:appendix-code}

The code implementation of FrameRouter is publicly available at
\url{https://github.com/bingjunluo/FrameRouter}. The code is organized to match the method contract in
Section~\ref{sec:method:framework}: the router first reads only the current
question and its answer options, assigns one frame budget, and then runs
exactly one selected full-video uniform policy. It does not inspect video
frames, subtitles, labels, sibling questions, other videos, or any
candidate-policy outputs before route assignment. The released pipeline
therefore implements the deployable path used in the paper rather than a
post-hoc selector that runs multiple answer policies and chooses among their
outputs.

\paragraph{Module structure.}
The entry point is \texttt{run\_framerouter.py}, with command-line arguments
defined in \texttt{framerouter/cli.py}. Runtime and frame-budget settings are
kept in \texttt{framerouter/config.py}. The question-only routing prompt is in
\texttt{framerouter/router\_prompt.py}, while
\texttt{framerouter/router.py} performs the parser call, JSON extraction, and
parser TSV normalization. The deterministic fixed-mean budget assignment is
implemented in \texttt{framerouter/allocation.py}. After allocation,
\texttt{framerouter/policy.py} runs the selected 48/64/96-frame Video-MME-v2
inference policy, and \texttt{framerouter/scoring.py} wraps Video-MME-v2
scoring. The end-to-end orchestration lives in
\texttt{framerouter/pipeline.py}; \texttt{run\_videommev2.sh} provides an
example command for the main Video-MME-v2 setting.

\begin{center}
\begin{minipage}{0.86\linewidth}
\begin{Verbatim}[frame=single,framesep=2mm,fontsize=\small]
current question/options
  -> framerouter.router       # evidence-demand scores
  -> framerouter.allocation   # fixed-mean budget assignment
  -> framerouter.policy       # selected full-video policy only
  -> framerouter.scoring      # Video-MME-v2 metrics
\end{Verbatim}
\end{minipage}
\end{center}

\paragraph{Environment and execution.}
The paper experiments were run on an 8$\times$NVIDIA H200 server using vLLM
with the OpenAI-compatible chat-completions API and a Python 3.10+ client. The
inputs are Video-MME-v2 parquet metadata, extracted video files, and subtitle
JSONL files. Frames may be transported as base64-encoded JPEGs or through local
media paths supported by the inference server. The implementation is otherwise
model-server agnostic as long as the server accepts OpenAI-compatible
multimodal chat-completions requests.

A typical run first starts a vLLM server:
\begin{center}
\begin{minipage}{0.9\linewidth}
\begin{Verbatim}[frame=single,framesep=2mm,fontsize=\small]
python3 -m vllm.entrypoints.openai.api_server \
  --model <MODEL_PATH_OR_NAME> \
  --served-model-name qwen35-think \
  --tensor-parallel-size 8 \
  --port 8000 \
  --trust-remote-code
\end{Verbatim}
\end{minipage}
\end{center}
FrameRouter can then be launched with the provided script:
\begin{center}
\begin{minipage}{0.9\linewidth}
\begin{Verbatim}[frame=single,framesep=2mm,fontsize=\small]
bash run_videommev2.sh
\end{Verbatim}
\end{minipage}
\end{center}
For quick verification, the release also supports a small smoke test via
\texttt{--limit} and an allocation-only mode via \texttt{--only-allocate} when
an existing parser TSV is available.

\paragraph{Outputs.}
The script writes \texttt{parser.tsv} for question-only parser scores,
\texttt{routes.tsv} for route assignments and allocated frame counts,
\texttt{predictions.tsv} for selected-policy predictions and raw responses,
\texttt{predictions\_scored.tsv} for scored question-level outputs, and
\texttt{predictions\_rating.json} for aggregate Video-MME-v2 metrics.

\newpage

\end{document}